\documentclass{article}

\usepackage{mystyle}

\usepackage{microtype}
\usepackage{graphicx}
\usepackage{subfigure}
\usepackage{booktabs} 

\usepackage{hyperref}

\usepackage[linesnumbered, ruled, boxed]{algorithm2e}

\usepackage[accepted]{icml2020}

\icmltitlerunning{Ordinal Non-negative Matrix Factorization for Recommendation}

\begin{document}

\twocolumn[
\icmltitle{Ordinal Non-negative Matrix Factorization for Recommendation}



\icmlsetsymbol{equal}{*}

\begin{icmlauthorlist}
\icmlauthor{Olivier Gouvert}{cnrs}
\icmlauthor{Thomas Oberlin}{supaero}
\icmlauthor{Cédric Févotte}{cnrs}
\end{icmlauthorlist}

\icmlaffiliation{cnrs}{IRIT, Université de Toulouse, CNRS, France}
\icmlaffiliation{supaero}{ISAE-SUPAERO, Université de Toulouse, France}

\icmlcorrespondingauthor{Olivier Gouvert}{oliviergouvert@gmail.com}

\icmlkeywords{Machine Learning, ICML}

\vskip 0.3in
]



\printAffiliationsAndNotice{}  

\begin{abstract}
We introduce a new non-negative matrix factorization (NMF) method for ordinal data, called OrdNMF. Ordinal data are categorical data which exhibit a natural ordering between the categories. In particular, they can be found in recommender systems, either with explicit data (such as ratings) or implicit data (such as quantized play counts). OrdNMF is a probabilistic latent factor model that generalizes Bernoulli-Poisson factorization (BePoF) and Poisson factorization (PF) applied to binarized data. Contrary to these methods, OrdNMF circumvents binarization and can exploit a more informative representation of the data. We design an efficient variational algorithm based on a suitable model augmentation and related to variational PF. In particular, our algorithm preserves the scalability of PF and can be applied to huge sparse datasets. We report recommendation experiments on explicit and implicit datasets, and show that OrdNMF outperforms BePoF and PF applied to binarized data.
\end{abstract}

\section{Introduction}

Collaborative filtering (CF) is a popular recommendation technique based only on the feedbacks of users on items. These feedbacks can be stored into a matrix $\Y$ of size $U\times I$, where $U$ and $I$ are the number of users and items respectively. Matrix factorization (MF) methods~\cite{hu_collaborative_2008,koren_matrix_2009,ma_probabilistic_2011} aim to approximate the feedback matrix $\Y$ by a low-rank structure $\W\H^T$ where $\W\in\mathbb{R}^{U\times K}_+$ corresponds to user preferences and $\H\in\mathbb{R}^{I\times K}_+$ to item attributes.

Poisson factorization (PF)~\cite{canny_gap_2004,cemgil_bayesian_2009,gopalan_scalable_2015} is a non-negative matrix factorization (NMF) model~\cite{lee_learning_1999,lee_algorithms_2001,fevotte_algorithms_2011} which aims to predict future interactions between users and items in order to make recommendations. For this purpose, PF is often applied to a binarized version of the data, i.e., $\Y\in\{0,1\}^{U\times I}$, containing only the information that a user is interacting with an item or not. A variant of PF, called Bernoulli-Poisson factorization (BePoF)~\cite{acharya_nonparametric_2015}, has been proposed to explicitly model binary data. However, for both PF and BePoF, the binarization stage induces a loss of information, since the value associated to an interaction is removed. Although several attempts in the literature tried to directly model raw data, both for explicit~\cite{hernandez-lobato_probabilistic_2014} and implicit data~\cite{basbug_hierarchical_2016,zhou_nonparametric_2017,gouvert_recommendation_2019}, this remains a challenging problem.

{In an attempt to keep as much information as possible, we propose in this paper to consider ordinal rather than binary data}. Ordinal data~\cite{stevens_theory_1946} are nominal/categorical data which exhibit a natural ordering (for example: cold $ \prec $ warm $ \prec $ hot). This type of data is encountered in recommender systems with explicit data such as ratings. {It can also be created by quantizing implicit data such as play counts. Such a pre-processing remains softer than binarization and stays closer to the raw data, as soon as the number of classes is chosen big enough}. In this paper, without loss of generality, we will work with ordinal data belonging to $\{0,\dots,V\}$.  Note that for this type of data, the notion of distance between the different classes is not defined. For example, this implies that the mean is not adapted to these data, unlike the median. 

{There are two naive ways to process ordinal data. The first one consists in applying classification methods. The scale of the ordering relation which links the different categories is then ignored. The second way considers these data as real values in order to apply regression models. By doing this, it artificially creates a distance between the different categories}. These two naive methods do not fully consider the specificity of ordinal data, since they remove or add information to the data. Threshold models~\cite{mccullagh_regression_1980,verwaeren_learning_2012} are {popular ordinal data processing methods} that alleviate this issue. They assume that the data results from the quantization of continuous latent variables with respect to (w.r.t.) an increasing sequence of thresholds. The aim of these models is then to train a predictive model on the latent variables and to learn the sequence of thresholds. Threshold models can thus be seen as an extension of naive regression models, where the distances between the different classes are learned through quantization thresholds. {For a comprehensive and more detailed review on ordinal regression methods, we refer to~\cite{gutierrez_ordinal_2015}.}

In this paper, we develop a new probabilistic NMF framework for ordinal data, called ordinal NMF (OrdNMF). OrdNMF is a threshold model where the latent variables have an NMF structure. {In other words, this amounts to defining the approximation $ \Y \approx G(\W \H^T)$, where $\Y$ is the ordinal data matrix, $\W$ and $\H$ are non-negative matrices and $G(\cdot)$ is a link function}. OrdNMF allows us to work on more informative class of data than classical PF method by circumventing binarization. 
Contrary to ordinal MF (OrdMF) models~\cite{chu_gaussian_2005,koren_ordrec_2011,paquet_hierarchical_2012,hernandez-lobato_probabilistic_2014}, OrdNMF imposes non-negativity constraints on both $\W$ and $\H$. This implies a more intuitive part-based representation of the data \cite{lee_learning_1999}, and were shown to improve results in recommendation \cite{gopalan_scalable_2015}.
OrdNMF can efficiently take advantage of the sparsity of $\Y$, scaling with the number of non-zero observations. Thereby, it can be applied to huge sparse datasets such as those commonly encountered in recommender systems.
As opposed to learning-to-rank models, the aim of OrdNMF is to model ordinal data, via a generative probabilistic model, in order to predict the class of future interactions. Learning-to-rank models do not seek to predict a class but to rank items relatively to each other. For example, Bayesian personalized ranking~\cite{BPR} is based on binary pairwise comparisons of the users' preferences and not on the raw matrix $\Y$. Although such models can also be used for recommendation, they are not generative.

The contributions of this paper are the following.

$\bullet$ We propose a new NMF model for ordinal data based on multiplicative noise. In particular, we study an instance of this model where the noise is assumed to be drawn from an inverse-gamma (IG) distribution. We show that this instance is an extension of BePoF~\cite{acharya_nonparametric_2015} and PF~\cite{gopalan_scalable_2015} applied to binarized data.

$\bullet$ We use a model augmentation trick to design an efficient variational algorithm, both for the update rules of the latent factors $\W$ and $\H$, and for those of the thresholds $\ve{b} $. In particular, this variational algorithm scales with the number of non-zero values in $\Y$.

$\bullet$ We report the results of OrdNMF on recommendation tasks for two datasets (with explicit and implicit feedbacks). Moreover, posterior predictive checks (PPCs) demonstrate the excellent flexibility of OrdNMF {and its ability to represent various kinds of datasets}.

The rest of the paper is organized as follows. In Section~\ref{sec:onmf prerequis}, we present {important related works} on cumulative link models and on BePoF. In Section~\ref{sec:onmf model}, we present our general OrdNMF model and detail a particular instance. In Section~\ref{sec:onmf inference}, we develop an efficient VI algorithm which scales with the number of non-zero values in the data. In Section~\ref{sec:onmf expe implcite}, we test our algorithm on recommendation tasks for explicit and implicit datasets. Finally, in Section~\ref{sec:onmf discussion}, we conclude and discuss the perspectives of this work.

\section{Related Works} \label{sec:onmf prerequis}

\subsection{Cumulative Link Models (CLMs)} \label{sec:onmf omf}

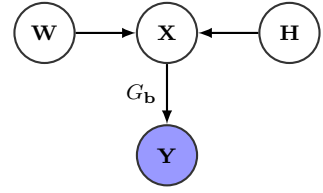
\begin{figure}[t]
\centering
\tikzset{
  font={\fontsize{8pt}{12}\selectfont}}
\begin{tikzpicture}
	\tikzstyle{main}=[circle, minimum size = 8mm, thick, draw =black!80, node distance = 8mm]
	\tikzstyle{connect}=[-latex, thick]
	\tikzstyle{box}=[rectangle, draw=black!100]

	\node[main] (w) [label=center:$\W$] { };
	\node[main] (c) [right=of w,label=center:$\X$] { };
	\node[main, fill = blue!40] (n) [below=of c,label=center:$\Y$] { };
	\node[main] (h) [right=of c,label=center:$\H$] { };

	\path (w) edge [connect] (c)
	    (h) edge [connect] (c);
	\path[-latex,thick] (c) edge node[anchor=center, left, midway] {$G_\ve{b}$} (n);
\end{tikzpicture}
\caption{Graphical model of OrdMF. A latent variable $\X$ is introduced to make the link between the factorization term $\W\H^T$ and the ordinal data $\Y$.}
\label{fig:onmf graph}
\end{figure}

CLMs were one of the first threshold models proposed for ordinal regression~\cite{agresti_categorical_2011}. These models have been adapted to deal with the MF problem, leading to OrdMF models. {They amount to finding the approximation $\Y\approx G(\W\H^T)$, where $\Y\in\{0,\dots,V\}^{U\times I}$ is an ordinal data matrix, $\W\in\mathbb{R}^{U\times K}$ and $\H\in\mathbb{R}^{I\times K}$ are latent factors, and $G(\cdot)$ is a parametrized link function described subsequently}. OrdMF has been applied mainly to explicit data in order to predict users feedbacks~\cite{chu_gaussian_2005,paquet_hierarchical_2012}.

\begin{figure}[t]
	\centering
	\includegraphics[height=5cm]{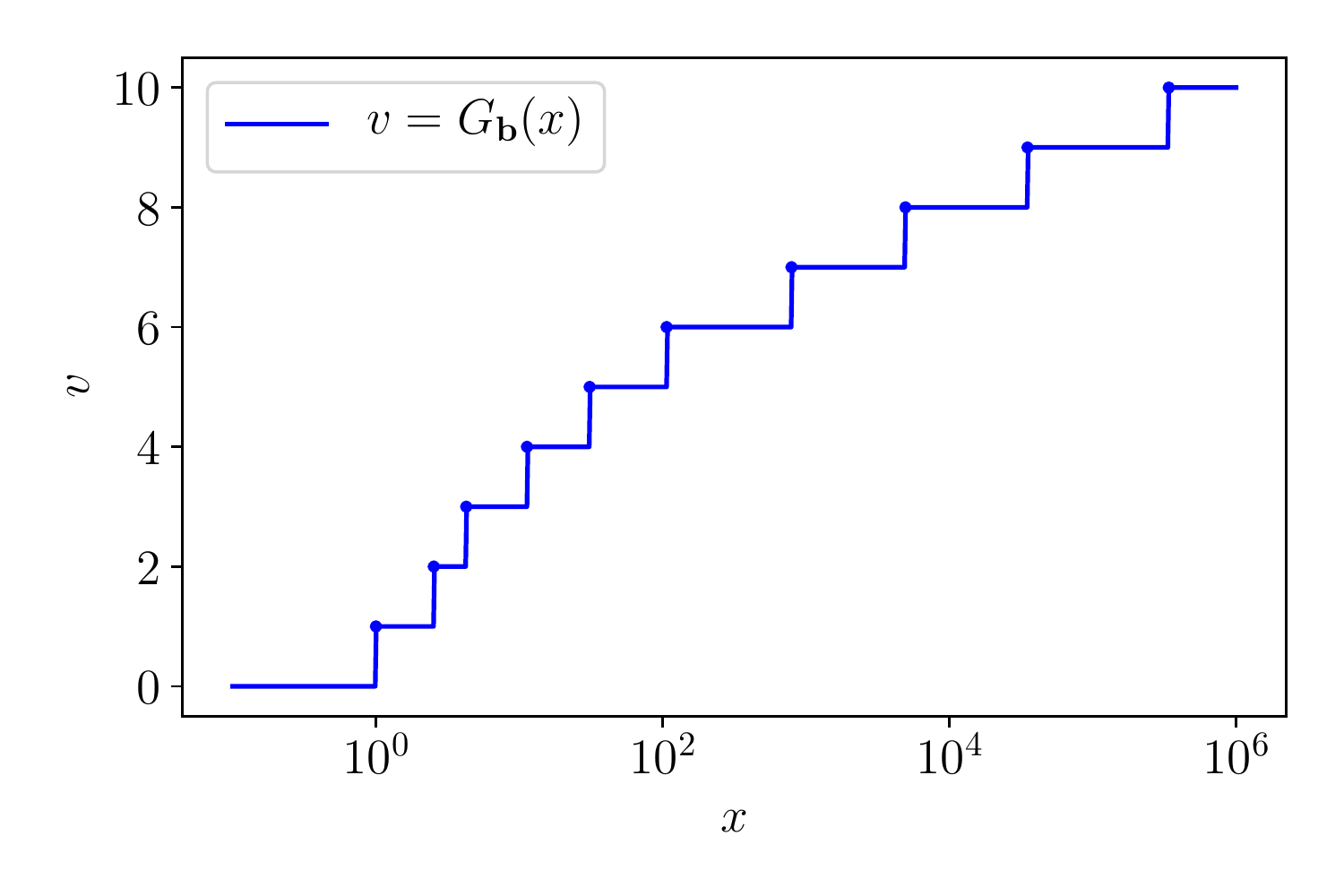}
	\vskip -0.2in
	\caption{Example of a quantization function $x \mapsto G_\ve{b}(x)$.}
	\label{fig:onmf quantif}
\end{figure}

The idea behind threshold models is to introduce a continuous latent variable $x_{ui}\in\mathbb{R}$ that is mapped to the ordinal data $y_{ui}$. {This is done by considering} an increasing sequence of thresholds $b_{-1}=-\infty < b_0<\dots<b_{V-1}<b_V=+\infty$, denoted by $\ve{b}$, {which fully characterize the following quantization function, illustrated in Figure~\ref{fig:onmf quantif}, by}:
\bal{
	\begin{array}{llcl}
	G_\ve{b}:&\mathbb{R} & \to & \{0,\dots,V\} \\
	&x & \mapsto & v \text{ such as } x\in[b_{v-1},b_v). \\
	\end{array} \label{eq:onmf quantif}
}

Therefore, ordinal data result from the quantization of the variable $x_{ui}$ by the step function $G_\ve{b}$, i.e., $y_{ui}=G_\ve{b}(x_{ui})$. The latent variable $x_{ui}$ corresponds to the variable $\lambda_{ui}=[\W\H^T]_{ui}\in\mathbb{R}$ perturbed by an additive noise $\varepsilon_{ui}$, whose cumulative density function (c.d.f.) is denoted by $F_\varepsilon:\mathbb{R}\to[0,1]$. Thus, we obtain the following generative model, illustrated in Figure~\ref{fig:onmf graph}:
\bal{
	&x_{ui}=\lambda_{ui} + \varepsilon_{ui}, \label{eq:cml1} \\
	&y_{ui}=G_\ve{b}(x_{ui}). \label{eq:cml2}
}

The goal of MF models for ordinal data is therefore to jointly infer the latent variables $\W$ and $\H$ as well as the sequence of thresholds $\ve{b}$.

\paragraph{Cumulative distribution function.}
The c.d.f. associated with the random variable $y_{ui}$ in Eqs.~\eqref{eq:cml1}-\eqref{eq:cml2} can be calculated as follows:
\bal{
	\mathbb{P}[y_{ui}\leq v|\lambda_{ui}] &=  \mathbb{P}[G_\ve{b}(x_{ui})\leq v|\lambda_{ui}] \\
	&= \mathbb{P}[\lambda_{ui} + \varepsilon_{ui}<b_v] \\
	&= \mathbb{P}[\varepsilon_{ui} < b_v - \lambda_{ui}] \\
	&= F_\varepsilon\left(b_v - \lambda_{ui}\right). \label{eq:onmf link mf}
}
It follows that the function $v\mapsto \mathbb{P}[y_{ui}\leq v|\lambda_{ui}]$ is increasing since the sequence of thresholds is itself increasing. Moreover, the probability mass function (p.m.f.) associated to the ordinal data can be written as:
\bal{
	\mathbb{P}[y_{ui}=v |\lambda_{ui}] 
	&= \mathbb{P}[y_{ui}\leq v|\lambda_{ui}] - \mathbb{P}[y_{ui}\leq v-1|\lambda_{ui}] \notag\\
	&= F_\varepsilon\left(b_v - \lambda_{ui}\right) - F_\varepsilon\left(b_{v-1} - \lambda_{ui}\right).
}

\paragraph{Some examples.}
If the c.d.f. is strictly increasing, we can rewrite Eq.~\eqref{eq:onmf link mf} as:
\bal{
	F_\varepsilon^{-1}(\mathbb{P}[y_{ui}\leq v|\lambda_{ui}]) = b_v - \lambda_{ui}.
}
Hence the name of CLM, since the factorization model is related to the c.d.f. of the ordinal data through a link function $F_\varepsilon^{-1}:[0,1]\to\mathbb{R}$. Various choices of noise (equivalently, of link function $F_\varepsilon^{-1}$) have been considered in the literature. We present some of these choices in what follows.

$\bullet$ Logit function. The use of the logit function was first proposed in~\cite{walker_estimation_1967}. This model was popularized and renamed as "proportional odds model" by~\cite{mccullagh_regression_1980}. The model can be rewritten as: 
\bal{
	\operatorname{logit} \mathbb{P}[y_{ui}\leq v|\lambda_{ui}] 
	= \log \dfrac{\mathbb{P}[y_{ui}\leq v|\lambda_{ui}]}{\mathbb{P}[y_{ui}> v|\lambda_{ui}]}
	= b_v - \lambda_{ui}
} 

$\bullet$ Probit function.
A common choice for the additive noise is $\varepsilon_{ui}\sim\mathcal{N}(0,\sigma^2)$~\cite{chu_gaussian_2005,paquet_hierarchical_2012,hernandez-lobato_probabilistic_2014}. In that case the link function $F_\varepsilon^{-1}$ is the probit function. Inference can be carried out with an EM algorithm based on the latent variable $x_{ui}$.

$\bullet$ Other choices like log-log or cauchit functions have also been considered~\cite{agresti_categorical_2011}. The survey~\cite{ananth_regression_1997} recaps some of these choices.

\subsection{Bernoulli-Poisson Factorization (BePoF)} \label{sec:onmf bepo}
In this section, we present BePoF~\cite{acharya_nonparametric_2015} which is a variant of PF for binary data (not directly related to the CLMs introduced above). It employs a model augmentation trick for inference that we will use in our own algorithm presented in Section~\ref{sec:onmf inference}.

The Poisson distribution can easily be ``augmented'' to fit binary data $y_{ui}\in\{0,1\}$. Indeed, it suffices to introduce a thresholding operation that binarizes the data. The corresponding generative hierarchical model is therefore given by:
\bal{
	&n_{ui} \sim \poisson([\W\H^T]_{ui}), \\
	&y_{ui} = \mathbb{1}[n_{ui}>0],
}
where $n_{ui}\in\mathbb{N}$ is a latent variable and $\mathbb{1}$ is the indicator function. We denote by $\N\in\mathbb{N}^{U\times I}$ the matrix such that $[\N]_{ui}=n_{ui}$. This variable can easily be marginalized by noting that ${\mathbb{P}[y_{ui}=0]= \poisson(0|[\W\H^T]_{ui})= e^{-[\W\H^T]_{ui}}}$. We obtain:
\bal{
	y_{ui} \sim \bernoulli(1-e^{-[\W\H^T]_{ui}}) \label{eq:onmf bepo}
}
where $\bernoulli$ refers to the Bernoulli distribution. The conditional distribution of the latent variable $n_{ui}$ is given by:
\bal{
	n_{ui}|y_{ui}\sim 
	\begin{cases}
	\delta_0, & \text{if } y_{ui}=0, \\
	\ZTP([\W\H^T]_{ui}), & \text{if } y_{ui}=1. \\
	\end{cases}
}
where $\ZTP$ refers to the zero-truncated Poisson distribution and $\delta_0$ to the Dirac distribution located in $0$. The latent variable $\N$ can be useful to design Gibbs or variational inference (VI) algorithms for binary PF~\cite{acharya_nonparametric_2015} and we will employ a similar trick in Section~\ref{sec:onmf inference}.

\paragraph{Remark.}
The generative model presented in Eq.~\eqref{eq:onmf bepo} is in the form $y_{ui} \sim \bernoulli(G([\W\H^T]_{ui}))$ where $G:\mathbb{R}~(\text{or }\mathbb{R}_+) \to [0,1]$. When $[\W\H^T]_{ui}\in\mathbb{R}$, the function $G$ can be the inverse of the probit~\cite{consonni_meanfield_2007} or of the logit function for example. They are special cases of the model presented in Section~\ref{sec:onmf omf} with $V=1$. Mean-parametrized Bernoulli MF models have also been considered~\cite{lumbreras_bayesian_2018}. They correspond to $G=\operatorname{Id}$ and require additional constraints on the latent factors $\W$ and $\H$ in order to satisfy $[\W\H^T]_{ui}\in [0,1]$.

\section{Ordinal NMF (OrdNMF)} \label{sec:onmf model}

In this section, we introduce OrdNMF which is a NMF model pecially designed for ordinal data. A difference with Section~\ref{sec:onmf omf} is that we impose that both matrices $\W$ and $\H$ are non-negative. Thus, we now have $[\W\H^T]_{ui}\in\mathbb{R}_+$ instead of $[\W\H^T]_{ui}\in\mathbb{R}$. We denote  $\lambda_{uik}=w_{uk}h_{ik}$ so that $\lambda_{ui}=\sum_k\lambda_{uik}=[\W\H^T]_{ui}$.

\subsection{Quantization of the Non-negative Numbers} \label{sec:onmf quantif reels}

Our model works on the same principle as OrdMF (see Section~\ref{sec:onmf omf}) and seeks to quantize the non-negative real line $\mathbb{R}_+$. For this, we introduce the increasing sequence of thresholds $\ve{b}$ given by $b_{-1}=0<b_0<\dots<b_{V-1}<b_V=+\infty$ (the thresholds are here non-negative). Moreover, we define the quantization function $G_\ve{b} : \mathbb{R}_+ \to \{0,\dots,V\}$ like in Eq.~\eqref{eq:onmf quantif} but with support $\mathbb{R}_+$.

As compared to Section~\ref{sec:onmf omf}, we now assume a non-negative multiplicative noise on $x_{ui}$. This ensures the non-negativity of $x_{ui}$ and {it seems well suited for modeling} over-dispersion, a common feature of recommendation data. Let $\varepsilon_{ui}$ be a non-negative random variable with c.d.f. $F_\varepsilon$, we thus propose the following generative model:
\bal{
	&x_{ui}=\lambda_{ui} \cdot \varepsilon_{ui}, \\
	&y_{ui}=G_\ve{b}(x_{ui}).
}

Like before, our goal is to jointly infer the latent variables $\W$ and $\H$ as well as the sequence of the thresholds $\ve{b}$. In our model, the c.d.f. associated to the ordinal random variable $y_{ui}$ becomes:
\bal{
	\mathbb{P}[y_{ui}\leq v|\lambda_{ui}] &=  \mathbb{P}[G_\ve{b}(x_{ui})\leq v|\lambda_{ui}] \\
	&= \mathbb{P}[\lambda_{ui}\cdot\varepsilon_{ui}<b_v] \\
	&= \mathbb{P}\left[\varepsilon_{ui} <\frac{b_v}{\lambda_{ui}}\right] \\
	&= F_\varepsilon\left(\frac{b_v}{\lambda_{ui}}\right). \label{eq:onmf fonction repartition}
}

Therefore, we can deduce that the p.m.f. is given by: 
\bal{
	&\mathbb{P}[y_{ui}=v |\lambda_{ui}] \notag\\
	&= \mathbb{P}[y_{ui}\leq v|\lambda_{ui}] - \mathbb{P}[y_{ui}\leq v-1|\lambda_{ui}] \\
	&= F_\varepsilon\left(\frac{b_v}{\lambda_{ui}}\right) - F_\varepsilon\left(\frac{b_{v-1}}{\lambda_{ui}}\right).
}

\begin{figure}[t]
		\centering
		\includegraphics[height=3cm]{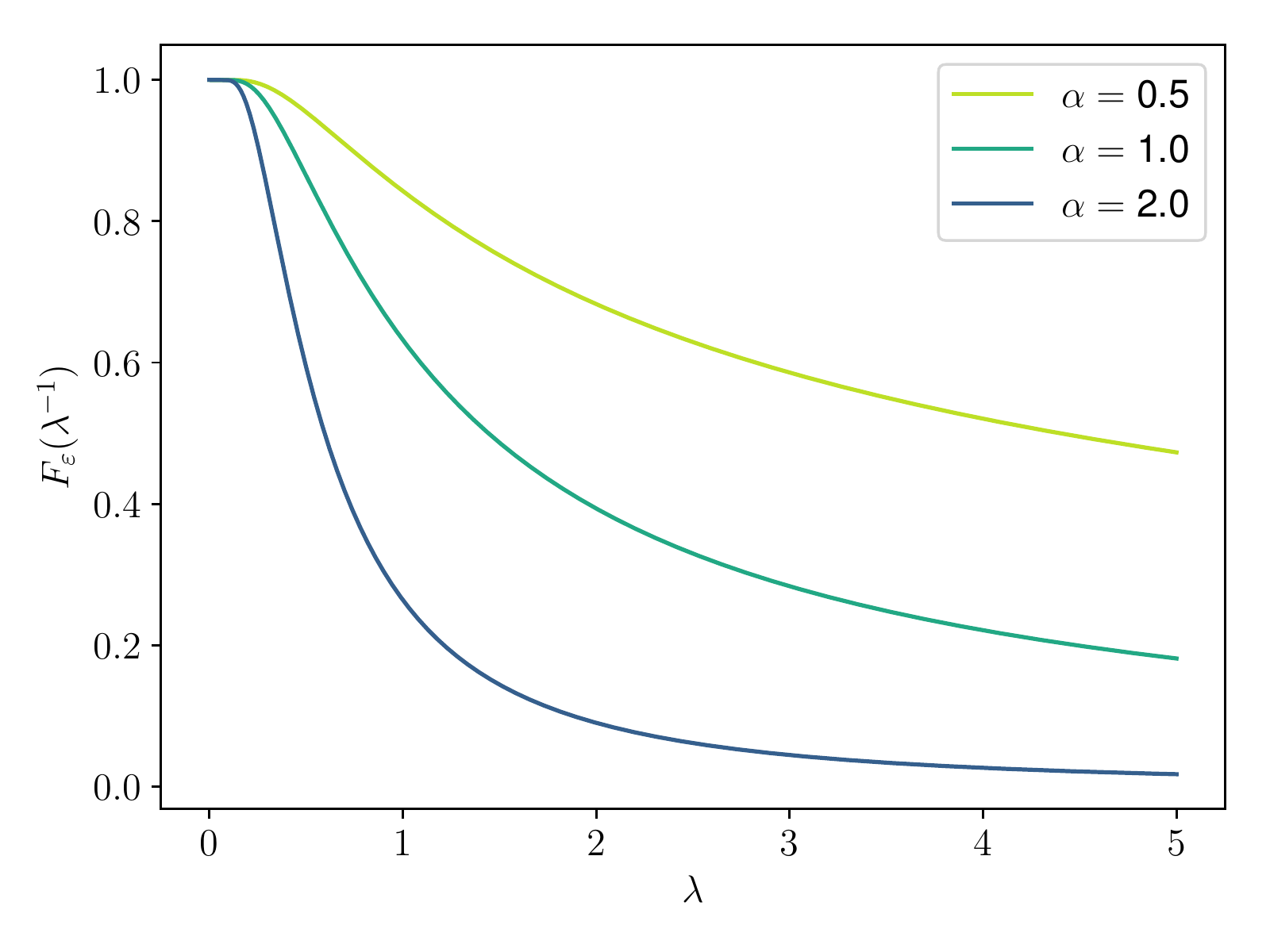}
		\includegraphics[height=3cm]{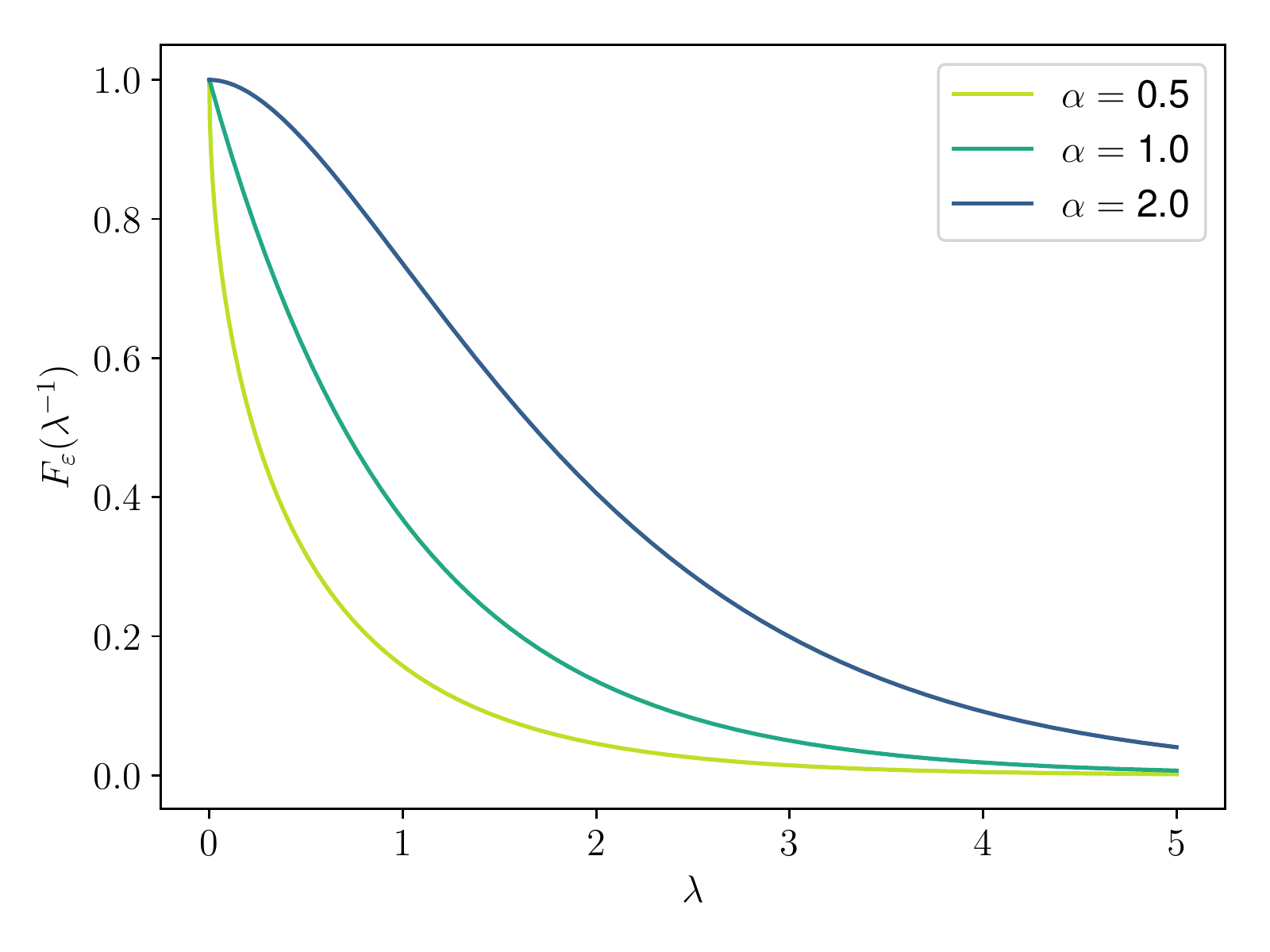}
		\caption{Functions $\lambda \mapsto F_\varepsilon(\lambda^{-1})$ for gamma (left) and inverse-gamma (right) noises.}
		\label{fig:onmf link function}
\end{figure}

Various functions $F_\varepsilon$ can be used {which determine the exact nature of the multiplicative noise}. Figure~\ref{fig:onmf link function} displays the function $\lambda \mapsto F_\varepsilon(\lambda^{-1})$ for the examples considered next.

$\bullet$ Gamma noise: $\varepsilon_{ui}\sim \dgamma(\alpha,1)$.\footnote{The rate parameter $\beta$ is fixed to $1$ because of a scale invariance with $\lambda_{ui}$.} The c.d.f. is given by $F_\varepsilon(x)=\frac{\gamma(\alpha,x)}{\Gamma(\alpha)}$ where $\gamma(\alpha,x)=\int_0^x t^{\alpha-1}e^{-t}\dint t$ is the lower incomplete gamma function. If $\alpha=1$, we recover an exponential noise $\varepsilon_{ui}\sim \Exp(1)$ whose c.d.f. is $F_\varepsilon(x)=1-e^{-x}$.

$\bullet$ Inverse-gamma (IG) noise: $\varepsilon_{ui}\sim \IG(\alpha,1)$.\addtocounter{footnote}{-1}\footnotemark{} 
The c.d.f. is given by $F_\varepsilon(x)=\frac{\Gamma(\alpha,x^{-1})}{\Gamma(\alpha)}$ where $\Gamma(\alpha,x)=\int_x^\infty t^{\alpha-1}e^{-t}\dint t$ is the upper incomplete gamma function. If $\alpha=1$, we obtain the c.d.f. $F_\varepsilon(x)= e^{-1/x}$.

$\bullet$ Any increasing function $F_\varepsilon:\mathbb{R}_+\to[0,1]$ defines a non-negative random variable which can be used in OrdNMF.

\subsection{OrdNMF with IG Noise (IG-OrdNMF)} \label{sec:ig-ordnmf}

In the rest of the paper, we focus on the special case where $\varepsilon_{ui}$ is a multiplicative IG noise with shape parameter $\alpha=1$, i.e., $\varepsilon_{ui}\sim \IG(1,1)$\footnote{The expectation of a IG variable is not defined for $\alpha\leq 1$, however the model is still well-defined.}. We use the acronym IG-OrdNMF for this particular instance of OrdNMF.

For convenience we write $\theta_v = b_v^{-1}$. The sequence $\boldsymbol\theta$ corresponds to the inverse of the thresholds and is therefore decreasing, i.e., $\theta_{-1}=+\infty>\theta_0>\dots>\theta_{V-1}>\theta_V=0$. Moreover, we denote by $\boldsymbol\Delta$ the {positive} sequence of decrements defined by $\Delta_v = \theta_{v-1}-\theta_v$ for $v\in\{1,\dots,V\}$. We have $\theta_v = \sum_{l=v+1}^V \Delta_l$ and, in particular, $\theta_{V-1}=\Delta_V$.

\paragraph{Interpretation.}
In IG-OrdNMF model, the c.d.f. associated with an ordinal data $y_{ui}$ is given by:
\bal{
	&\mathbb{P}[y_{ui}\leq v|\lambda_{ui}] = e^{-\lambda_{ui}\theta_{v}}, \\
	\text{or } &\mathbb{P}[y_{ui} > v|\lambda_{ui}] = 1-e^{-\lambda_{ui}\theta_{v}},
}
with $v\in\{0,\dots,V\}$. Therefore, BePoF (see Section~\ref{sec:onmf bepo}) is a particular case of IG-OrdNMF with $V=1$ and $\theta_0=1$.

This formulation allows for a new interpretation of IG-OrdNMF. As a matter of fact, the event $\{y_{ui}> v\}$ is a binary random variable which follows a Bernoulli distribution: $\{y_{ui}> v\}\sim \bernoulli(1-e^{-\lambda_{ui}\theta_{v}})$. Then, we can see IG-OrdNMF as the aggregation of $V$ dependent BePoF models for different thresholds of binarization $v\in\{0,\dots,V-1\}$.

\paragraph{Probability mass function.}
The p.m.f. of an observation is given by:
\bal{
	\mathbb{P}[y_{ui}=v|\lambda_{ui}] = 
	\begin{cases}
	e^{-\lambda_{ui}\theta_{0}}, & \text{for } v=0, \\
	e^{-\lambda_{ui}\theta_{v}} - e^{-\lambda_{ui}\theta_{v-1}}, & \text{for } 1\leq v < V, \\
	1 - e^{-\lambda_{ui}\theta_{V-1}}, & \text{for } v=V.
	\end{cases} 
}
Then, the log-likelihood of $\lambda_{ui}$ can be written as:
\bal{
	\log\mathbb{P}[y_{ui}=v|\lambda_{ui}] = 
	\begin{cases}
	-\lambda_{ui}\theta_{0}, \text{ if } v=0, \\
	-\lambda_{ui}\theta_{v} + \log (1 - e^{-\lambda_{ui}\Delta_v}), \text{ else}.
	\end{cases} \label{eq:onmf non lineaire}
}
This expression brings up a linear term in $\lambda_{ui}$ and a non-linear term of the form $x\mapsto\log(1-e^{-x})$, similar to the function used in Section~\ref{sec:onmf bepo}.

Moreover, the expectation of the observations is well-defined and given by :
\bal{
	\E{y_{ui}|\lambda_{ui}}=V-\sum_{v=0}^{V-1}e^{-\lambda_{ui}\theta_v}.
}
Note that, in the context of ordinal data processing, the expectation is not a good statistic since it implicitly implies a notion of distance between classes. However, this quantity will be useful to build lists of recommendations. Indeed, the function  $\lambda_{ui}\mapsto\E{y_{ui}|\lambda_{ui}}$ is increasing. Thus, the higher the $\lambda_{ui} = [\W \H^T]_{ui}$, the higher (in expectation) the level of interaction between the user and the item. 

\section{Bayesian Inference} \label{sec:onmf inference}

We impose a gamma prior on the entries of both matrices $\W$ and $\H$, i.e., $w_{uk}\sim\dgamma(\alpha^W,\beta^W_u)$ and $h_{ik}\sim\dgamma(\alpha^H,\beta^H_i)$. Gamma prior is known to induce sparsity which is a desirable property in NMF methods.

\subsection{Augmented Model} \label{sec:onmf augmentation}

As described in Section~\ref{sec:ig-ordnmf} , the log-likelihood for ordinal data such that $v\in\{1,\dots,V\}$ brings up a non-linear term $\log(1-e^{-x})$ which is not conjugate with the gamma distribution, making the inference complicated. To solve this issue, we use the trick presented in Section~\ref{sec:onmf bepo} by augmenting our model with the latent variable:
\bal{
	n_{ui}|y_{ui},\lambda_{ui}\sim 
	\begin{cases}
	\delta_0, & \text{if } y_{ui}=0, \\
	\ZTP(\lambda_{ui}\Delta_{y_{ui}}), & \text{if } y_{ui}>0. \\
	\end{cases}
} 
Moreover, as commonly done in the PF setting~\cite{cemgil_bayesian_2009,gopalan_scalable_2015}, we augment our model with the latent variable $\ve{c}_{ui}|n_{ui},\lambda_{ui} \sim \Mult(n_{ui},\boldsymbol\phi_{ui})$, where $\Mult$ is the multinomial distribution and $\boldsymbol\phi_{ui}$ is a probability vector with entries $\frac{\lambda_{uik}}{\lambda_{ui}}$.
Therefore, for ordinal data $y_{ui}\in\{1,\dots,V\}$, we obtain the following joint log-likelihood:
\bal{
	&\log p(y_{ui},n_{ui},\ve{c}_{ui}|\lambda_{ui}) 
	= -\lambda_{ui}\theta_{{y_{ui}}-1} \\
	&+ n_{ui}\log\Delta_{y_{ui}} + \sum_k \left( c_{uik}\log\lambda_{uik} - \log c_{uik}! \right), \notag \\
	& \text{ s.t. } n_{ui}\in\mathbb{N}^\ast \text{ and } n_{ui}=\sum_k c_{uik}. \notag
} 

\paragraph{Joint log-likelihood of IG-OrdNMF.}
We denote by $\Z=\{\N,\C,\W,\H\}$ the set of latent variables of the augmented model.
Moreover, we define $T_v$ such that:
\bal{
	T_v = 
	\begin{cases}
		\theta_0, & \text{if } v = 0, \\
		\theta_{v-1}, & \text{if } v > 0.
	\end{cases} \label{eq:onmf T}
 }
The joint log-likelihood of IG-OrdNMF is therefore given by:
\bal{
	&\log p (\Y,\N,\C|\W,\H)= 
	\sum_{ui} 
	\Big[ n_{ui}\log\Delta_{y_{ui}} \notag\\
	& + \sum_k \left( c_{uik}\log\lambda_{uik} - \log c_{uik}! \right) 
	- \lambda_{ui} T_{y_{ui}} \Big].
}

It is important to note that $n_{ui}=0$ and $\ve{c}_{ui}=\ve{0}_K$ when $y_{ui}=0$. Consequently, the variables $\N$ and $\C$ are partially observed and the inference take advantage of the sparsity of the observed matrix $\Y$.

\subsection{Variational Inference} \label{sec:onmf vi}

\begin{table}[t]
\caption{Variational distributions for IG-OrdNMF.}
\vskip 0.15in
\centering
\begin{tabular}{cl}
\toprule
Var. & Distribution \\
\midrule
$\C$ & $q(\ve{c}_{ui}|n_{ui}) = \mult\left(\ve{c}_{ui};~n_{ui},~\tilde{\boldsymbol\phi}_{ui}\right)$ \\

$\N$ & $q(n_{ui}) = 
\begin{cases}
\delta_0,& \text{ if } y_{ui}=0 \\
\ZTP(n_{ui};~\Lambda_{ui}\Delta_{y_{ui}}),& \text{ if } y_{ui}>0 
\end{cases}$ \\
$\W$ & $q(w_{uk})=\dgamma\left(w_{uk};~\tilde\alpha^W_{uk},~\tilde\beta^W_{uk}\right)$ \\ 

$\H$ & $q(h_{ik})=\dgamma\left(h_{ik};~\tilde\alpha^H_{ik},~\tilde\beta^H_{ik}\right)$ \\
\bottomrule
\end{tabular}
\label{tab:oNMF VI}
\end{table}

The posterior distribution $p(\Z|\Y)$ is intractable. We use VI to approximate this distribution by a simpler variational distribution $q$. Here, we assume that $q$ belongs to the mean-field family and can be written in the following factorized form:
\bal{
	q(\Z) = \prod_{ui}q(n_{ui},\ve{c}_{ui})\prod_{uk}q(w_{uk})\prod_{ik}q(h_{ik}).
}
Note that the variables $\N$ and $\C$ remains coupled. We use a coordinate-ascent VI (CAVI) algorithm to optimize the parameters of $q$. The variational distributions are described in Table~\ref{tab:oNMF VI}. The associated update rules are summarized in Algorithm~\ref{algo:onmf implicit}. 

\paragraph{Approximation and link with PF.}
Algorithm~\ref{algo:onmf implicit} can be simplified by assuming that $q(n_{ui})=\delta_1$ if $y_{ui}>0$. This amounts to replacing the non-linear term $\log (1-e^{-x})$ by $\log x$ in Eq.~\eqref{eq:onmf non lineaire}. However, this approximation will produce similar results only if $x$ is very small, since $\log (1-e^{-x}) = \log x + o(x)$. In practice, this can only be verified a posteriori by observing that $\Eq{n_{ui}}\approx 1$.

As mentioned above, BePoF is a special case of IG-OrdNMF for $V=1$ and $\theta_0=1$. Thus, we can notice that PF algorithm applied to binary data is an approximation of BePoF algorithm for $q(n_{ui})=\delta_1$ if $y_{ui}=1$. 

\subsection{Thresholds Estimation}
A key element of threshold models is the learning of thresholds (corresponding here to $\boldsymbol\theta$ parameters). For this, we use a VBEM algorithm. It aims to maximize the term $\Eq{\log p(\Y,\Z;\boldsymbol\theta)}$, w.r.t. the variables~$\boldsymbol\theta$, which is given by:
\bal{\label{eq:onmf optim theta}
	&\Eq{\log p(\Y,\Z;\boldsymbol\theta)} = \\
	&\sum_{ui} \Big[ \Eq{n_{ui}} \log\Delta_{y_{ui}} - \Eq{\lambda_{ui}} T_{y_{ui}} \Big]+ cst, \notag \\
	&\text{s.t. } \theta_0>\theta_1>\dots>\theta_{V-1}>\theta_V=0. \notag 
}
Note that both terms $T_v$ (defined in Eq.~(\ref{eq:onmf T})) and $\Delta_{v}=\theta_{v-1}-\theta_v>0$ depend on the sequence~$\boldsymbol\theta$.

\paragraph{Decrements optimization.}
We choose to work on the decrement sequence~$\boldsymbol\Delta$ rather than on the threshold sequence~$\boldsymbol\theta$. Indeed, by doing so, the decreasing constraint of~$\boldsymbol\theta$ becomes a non-negativity constraint of~$\boldsymbol\Delta$. Moreover, we obtain only terms in $x$ and $\log x$ in the function to be maximized. Thus, the problem can be solved analytically.

We can rewrite the term $T_v$ w.r.t. the sequence $\boldsymbol\Delta$ by noting that:
$T_v = \sum_{l=1}^{V} \mathbb{1}[v \leq l] \Delta_l,\, \forall v \in\{0,\dots,V\}$.
Therefore, the optimization problem presented in Eq.~(\ref{eq:onmf optim theta}) amounts to maximizing the following function:
\bal{
	&\Eq{\log p(\Y,\Z;\boldsymbol\Delta)}= 
	\sum_{ui}\sum_{l=1}^V 
	\Big[
	\mathbb{1}[y_{ui}=l] \Eq{n_{ui}} \log\Delta_l \notag \\
	& - \mathbb{1}[y_{ui} \leq l] \Eq{\lambda_{ui}} \Delta_l \Big]
	+ cst, 
	\text{ s.t. } \boldsymbol\Delta\geq0.
}

Thus, we obtain the following update rules:
\bal{
	&\Delta_l = \dfrac{\sum_{ui} \mathbb{1}[y_{ui}=l] \Eq{n_{ui}}}
	{\sum_{ui}\mathbb{1}[y_{ui} \leq l] \Eq{\lambda_{ui}} },
	\forall l\in\{1,\dots,V\}, \label{eq:onmf update delta} \\
	&\theta_v = \sum_{l=v+1}^V \Delta_l,
	\forall v\in\{0,\dots,V-1\}. \label{eq:onmf update theta}
}

\begin{algorithm}[t]
	\SetAlgoLined
	\KwData{Matrix $\Y$}
	\KwResult{Variational distribution $q$ and thresholds $\boldsymbol\theta$}
	Initialization of variational parameters and thresholds $\boldsymbol\theta$;
	\Repeat{ELBO converge}{
		\ForEach{couple $(u,i)$ such as $y_{ui}>0$}{
			~~$\Lambda_{uik} = \exp\left(\Eq{\log w_{uk}} + \Eq{\log h_{ik}}\right)$; \\ 
			$\Lambda_{ui}=\sum_k \Lambda_{uik}$;\\
			$\Eq{n_{ui}} = \frac{\Lambda_{ui}\Delta_{y_{ui}}}{1-e^{-\Lambda_{ui}\Delta_{y_{ui}}}};$ \\
			$\Eq{c_{uik}} = \Eq{n_{ui}} \frac{\Lambda_{uik}}{\Lambda_{ui}}$;
		}
		\ForEach{user $u\in\{1,\dots,U\}$}{
			~~$\tilde{\alpha}^W_{uk} = \alpha^W+\sum_i \Eq{c_{uik}}$;\\
			$\tilde{\beta}^W_{uk} = \beta^W_u+ \sum_i T_{y_{ui}}\Eq{h_{ik}}$;  
		}
		\ForEach{item $i\in\{1,\dots,I\}$}{
			~~$\tilde{\alpha}^H_{ik} = \alpha^H+\sum_u \Eq{c_{uik}}$;\\
			$\tilde{\beta}^H_{ik} = \beta^H_i+ \sum_u T_{y_{ui}}\Eq{w_{uk}}$;  
		}
		~~Update of thresholds: Eq.~\eqref{eq:onmf update delta} and Eq.~\eqref{eq:onmf update theta}; \\
		Update of rate parameters $\beta^W_u$ and $\beta^H_i$; \\
		Calculate $\ELBO(q,\boldsymbol\theta)$;
	}
	\caption{CAVI for IG-OrdNMF.}
	\label{algo:onmf implicit}
\end{algorithm}

Algorithm~\ref{algo:onmf implicit} scales with the number of non-zero values in the observation matrix $\Y$. The complexity of OrdNMF is of the same order of magnitude as BePoF and PF. The only difference with these algorithms in terms of computational complexity is the update of the thresholds (Line 11 of Alg.~\ref{algo:onmf implicit}).

\begin{table*} \small
\centering

\caption{Recommendation performance of OrdNMF using the MovieLens dataset. Bold: best NDCG score. R: raw data. B: binarized data.}
\vskip 0.15in
\begin{tabular}{llllllllc}
	\toprule
	&  &  & \multicolumn{5}{c}{$\operatorname{NDCG}@100$ with threshold $s$} \\
	Model & Data & K & $s=1$ & $s=4$ & $s=6$ & $s=8$ & $s=10$  \\

	\midrule
	OrdNMF & R & $150$ & $\bf 0.444$ & $\bf 0.444$ & $\bf 0.439$ & $\bf 0.414$ & $0.353$ \\

	BePoF & B ($\geq1$) & $50$ & $0.433$ & $0.430$ & $0.421$ & $0.383$ & $0.310$ \\

	PF & B ($\geq1$) & $100$ & $0.431$ & $0.428$ & $0.418$ & $0.380$ & $0.306$ \\

	BePoF & B ($\geq 8$) & $50$ & $0.389$ & $0.393$ & $0.399$ & $0.408$ & $\bf 0.369$ \\

	PF & B ($\geq 8$) & $150$ &	$0.386$ & $0.389$ & $0.395$ & $0.403$ & $ 0.365$ \\

	\bottomrule
\end{tabular}
\label{tab:onmf resultats ml}
\end{table*}

\begin{table*} \small
\caption{Recommendation performance of OrdNMF using the Taste Profile dataset. Bold: best NDCG or log-likelihood score. Q: quantized data. B: binarized data. R: raw data.}
\centering
\vskip 0.15in
\begin{tabular}{lllllllllc}
	\toprule
	&  &  & \multicolumn{6}{c}{$\operatorname{NDCG}@100$ with threshold $s$} \\
	Model & Data & K & $s=1$ & $s=3$ & $s=6$ & $s=11$ & $s=21$ & $s=51$ & log-lik\\

	\midrule
	OrdNMF & Q & $250$ & \bf 0.213 & \bf 0.174 & 0.153 & 0.135 & 0.123 & 0.117 & $\bf -2.8 \cdot 10^{5}$ \\
	
	dcPF & R & $150$ & 0.209 & 0.173 &	\bf 0.154 & \bf 0.137 & \bf 0.128 & \bf 0.121 & $-3.0 \cdot 10^{5}$ \\

	BePoF & B ($\geq 1$) & $250$ & 0.210 & 0.170 & 0.149 & 0.131 & 0.120 & 0.115 & N/A \\

	PF & B ($\geq 1$) & $250$ & 0.206 & 0.167 & 0.146 & 0.129 & 0.118 & 0.115 & N/A \\
	\bottomrule
\end{tabular}
\label{tab:onmf resultats tps}
\end{table*}

\subsection{Posterior Predictive Expectation.} \label{sec:ppe}

The posterior predictive expectation $\E{\Y^\ast|\Y}$ corresponds to the expectation of the distribution of new observations $\Y^\ast$ given previously observed data $\Y$. This quantity allows us to create the list of recommendations for each user. We can approximate it by using the variational distribution $q$:
\bal{
	\E{\Y^\ast|\Y} \approx \int_{\W,\H}\E{\Y^\ast|\W,\H}q(\W)q(\H)\dint\W\dint\H. \label{eq:onmf approx esp pred}
}
Unfortunately, this expression is not tractable. {But for recommendation we are only interested in ordering items w.r.t. this quantity. The function $\lambda_{ui}\mapsto \E{y^\ast_{ui}|\lambda_{ui}}$ being increasing, we can use instead of Eq.~\eqref{eq:onmf approx esp pred} the simpler score} $s_{ui}=[\Eq{\W}\Eq{\H}^T]_{ui}$.

\section{Experimental Results} \label{sec:onmf expe implcite}

\subsection{Experimental Set Up}

\paragraph{Datasets.}

We report experimental results for two datasets described below.

\begin{figure*}[t]
	\centering
	\includegraphics[height=5.cm]{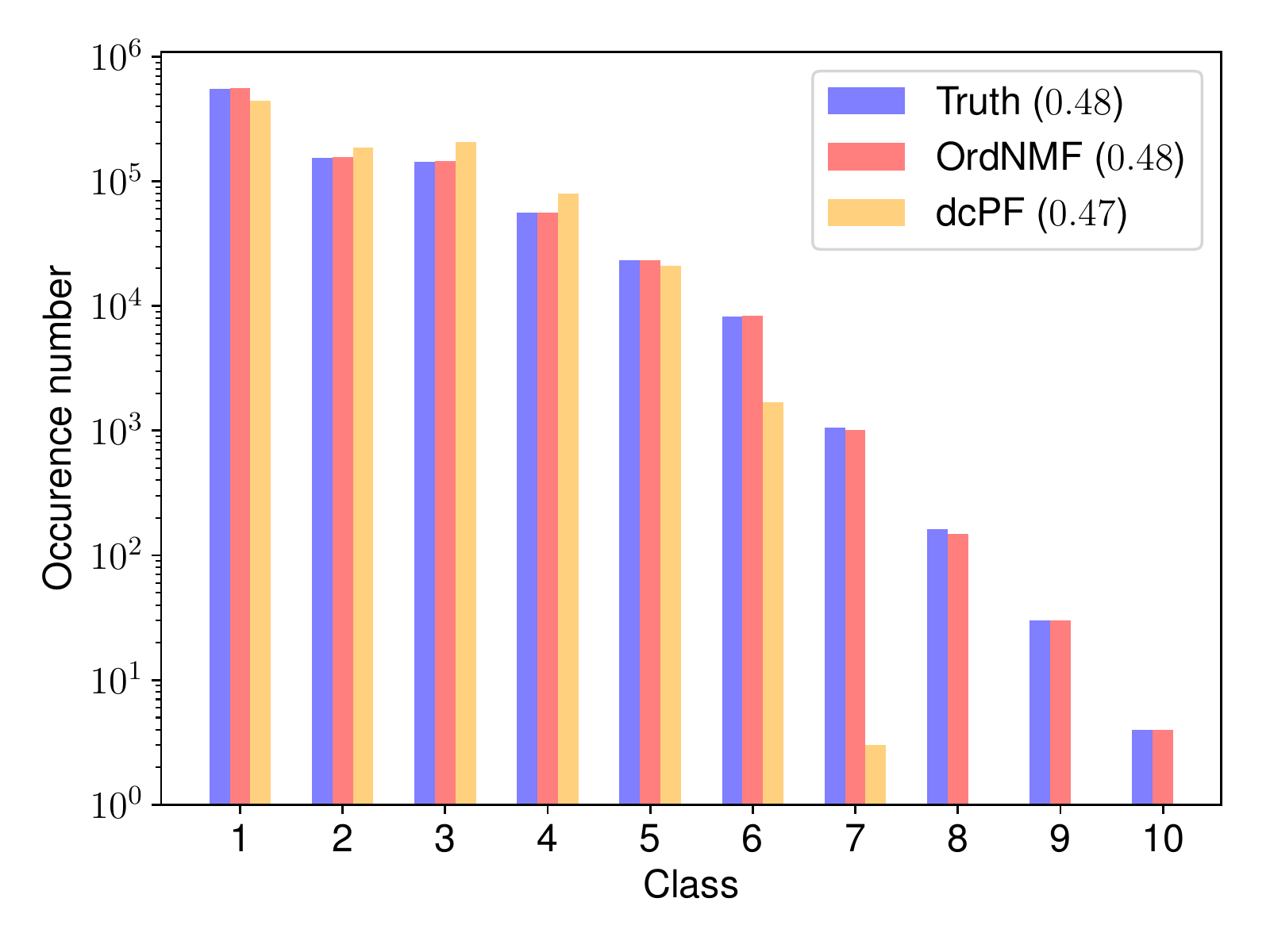}
	\hspace{1cm}
	\includegraphics[height=5.cm]{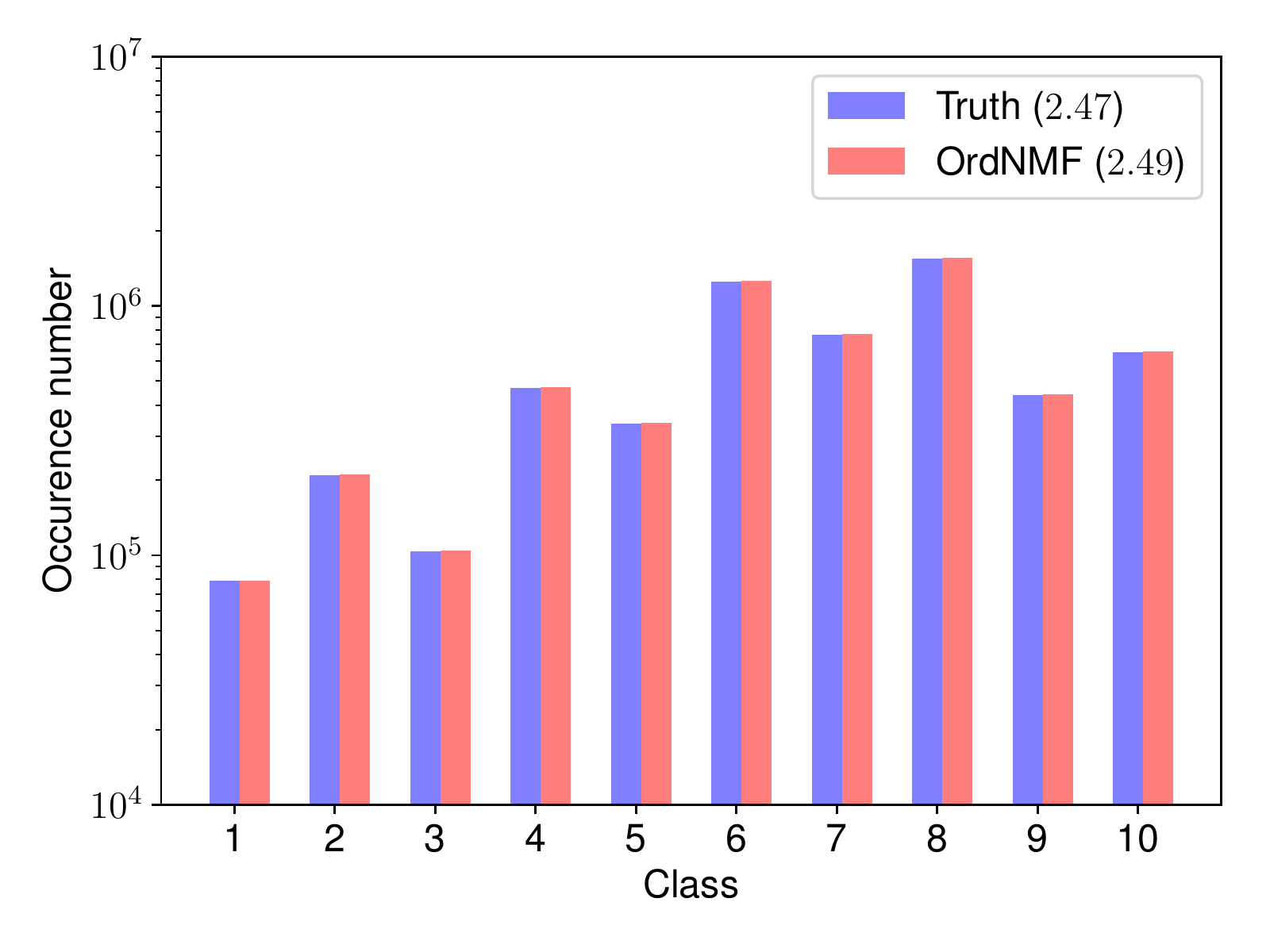}
	\vskip -0.1in
	\caption{PPC of the distribution of the classes in the Taste Profile dataset (left) and MovieLens dataset (right). The blue bars (Truth) represents the histogram of the classes in the train set. The colored bars represent the simulated histograms obtained from the different inferred OrdNMF or dcPF models. The percentages of non-zero values are written in parentheses.}
	\label{fig:onmf ppc}
\end{figure*}

$\bullet$ MovieLens~\cite{harper_movielens_2015}. This dataset contains the ratings of users on movies on a scale from $1$ to $10$. These explicit feedbacks correspond to ordinal data. We consider that the class $0$ corresponds to the absence of a rating for a couple user-movie. The histogram of the ordinal data is represented in blue on Figure~\ref{fig:onmf ppc}. We pre-process a subset of the data as in \cite{liang_modeling_2016}, keeping only users and movies that have more than 20 interactions. We obtain $U=20$k users and $I=12$k movies.

$\bullet$ Taste Profile~\cite{bertin-mahieux_million_2011}. This dataset, provided by the Echo Nest, contains the play counts of users on a catalog of songs. As mentioned in the introduction, we choose to quantize these counts on a predefined scale in order to obtain ordinal data. We {}arbitrarily select the following quantization thresholds: $[1,2,5,10,20,50,100,200,500]$. For example, the class labeled $6$ corresponds to a listening counts between $21$ and $50$. As for MovieLens, the class $0$ corresponds to users who have not listen to a song. The histogram of the ordinal data are displayed in blue on Figure~\ref{fig:onmf ppc}. We pre-process a subset of the data as before and obtain $U=16$k users and $I=12$k songs.

Be careful not to confuse the predefined quantization used to obtain ordinal data, with the quantization of the latent variable in OrdNMF model which is estimated during inference. {Although we expect OrdNMF to recover a relevant scaling between the categories, there is no reason to get the same quantization function that was used for pre-processing.}

\paragraph{Evaluation.} 
Each dataset is split into a train set $\Y^\text{train}$ and a test set $\Y^\text{test}$: the train set contains $80\%$ of the non-zero values of the original dataset $\Y$, the other values are set to the class $0$; the test set contains the remaining $20\%$. All the compared methods are trained on the train set and then evaluated on the test set. 

First, we evaluate the recommendations with a ranking metric.  For each user, we propose a list of $m=100$ items (movies or songs) ordered w.r.t. the prediction score presented in Section~\ref{sec:ppe}: $s_{ui}=[\Eq{\W}\Eq{\H^T}]_{ui}$. The quality of these lists is then measured through the NDCG metric~\cite{jarvelin_cumulated_2002}. The NDCG rewards relevant items placed at the top of the list more strongly than those placed at the end. We use the relevance definition proposed in~\cite{gouvert_recommendation_2019}:
\bal{
	\rel(u,i) = \mathbb{1}[y_{ui}^\text{test}\geq s].
}
In other words, an item is considered as relevant if it belongs at least to the class $s$ in the test set. The NDCG metric is between $0$ and $1$, the higher the better.

Moreover, for the Taste Profile dataset, we calculate the log-likelihood of the non-zero entries on the test set, as it is done in~\cite{basbug_hierarchical_2016}:
\bal{
	\mathcal{L}_{NZ} = \sum_{(u,i)\in\text{Testset}} \log p(y_{ui}^\text{test}|y_{ui}^\text{test}>0,\hat\W,\hat\H),
}
where $y_{ui}^\text{test}$ is the quantized data, $\hat w_{uk}=\mathbb{E}[w_{uk}]$ and $\hat h_{ik}=\mathbb{E}[h_{ik}]$ are the estimated latent factors. 

\paragraph{Compared methods.}
We compare OrdNMF with three other models: PF, BePoF and discrete compound PF (dcPF)~\cite{basbug_hierarchical_2016} with a logarithmic element distribution as implemented in~\cite{gouvert_recommendation_2019}. Each model is applied either to raw data (R), quantized data (Q) or binarized data (B). For the MovieLens dataset, two different binarizations are tested: one with a threshold at 1 ($\geq 1$) and one with a threshold at 8 ($\geq 8$). For the Taste Profile dataset, dcPF is applied to the count data (R) whereas OrdNMF is applied to the quantized data (Q).

For all models, we select the shape hyperparameters $\alpha^W=\alpha^H=0.3$ among $\{0.1,0.3,1\}$~\cite{gopalan_scalable_2015}. The number of latent factors is chosen among $K\in\{25,50,100,150,200,350\}$ for the best NDCG score with threshold $s=8$ for the MovieLens dataset, and $s=1$ for the Taste Profile dataset. All the algorithms are run 5 times with random initializations and are stopped when the relative increment of the expected lower bound (ELBO) falls under $\tau = 10^{-5}$. The computer used for these experiments was a MacBook Pro with an Intel Core i5 processor (2,9 GHz) and 16 Go RAM. All the Python codes are available on \url{https://github.com/Oligou/OrdNMF}.

\subsection{Prediction Results} \label{sec:onmf results tps}
Table~\ref{tab:onmf resultats ml} displays the results for the MovieLens dataset. First, we can compare BePoF with its approximation, i.e., PF applied to binarized data. BePoF is slightly better than PF for both binarizations, and requires less latent factors. Then, we observe that the choice of the binarization has a big impact on the NDCG scores. BePoF with data thresholded at 1 ($\geq1$) perform well on small NDCG threshold $s$ but has poor performance after. On the contray, with data thresholded at 8 ($\geq8$), BePoF achieves best performances for NDCG $s=10$ but poor performances for small $s$. OrdNMF does not exhibit such differences between NDCG scores and benefits from the additional information brought by the ordinal classes. Nevertheless, we can note a small decrease of the performance with $s=10$ which is the hardest class to predict.

Table~\ref{tab:onmf resultats tps} displays the same kind of results for the Taste Profile dataset. Again, OrdNMF outperforms BePoF and PF which exploit less data information. OrdNMF is competitive with dcPF which gives the best results for the highest thresholds $s$. However, OrdNMF presents a higher log-likelihood score than dcPF. Thus, OrdNMF seems better suited to predict the feedback class of a user than dcPF. This observation is confirmed by the posterior predictive checks (PPC) presented below.

\subsection{Posterior Predictive Check (PPC)} \label{sec:onmf ppc}

A PPC consists of generating new data based on the posterior predictive distribution $p(\Y^\ast,\W,\H|\Y)\approx p(\Y^\ast|\W,\H)q(\W)q(\H)$, and then compare the structure of the original data $\Y$ with the artificial data $\Y^\ast$. Here, we focus on the distribution of the ordinal categories. Figure~\ref{fig:onmf ppc} presents the results of these PPCs for the Taste Profile dataset. The blue bars correspond to the empirical histogram of the data ($\Y^\text{train}$), the red and orange bars correspond to the histograms of the simulated data obtained with OrdNMF and dcPF respectively. While dcPF fails to model the very large values present in the data (from the class $7$, which corresponds to values greater than $50$ plays), OrdNMF seems to precisely describe all the ordinal categories. This is also the case on the MovieLens dataset too. Even if the empirical histogram is here less regular, OrdNMF can adapt itself to all type of data through the inferred thresholds~$\ve{b}$.

\section{Conclusion} \label{sec:onmf discussion}

We developed a new probabilistic NMF framework to process ordinal data. In particular, we presented IG-OrdNMF which is an extension of BePoF and conducted experiments on two different datasets. We show the ability of OrdNMF to process different kinds of ordinal data both explicit and implicit. This work opens up several exciting perspectives. As we described in Section~\ref{sec:onmf quantif reels}, OrdNMF can be used for different choices of multiplicative noise. It would be interesting to develop OrdNMF for the exponential noise in a similar way than IG-OrdNMF. {Finally, when applied to implicit data, it would be of particular interest to learn the pre-processing during the factorization, in order to automatically tune the level of pre-processing adapted to a given dataset. This is yet left for future investigations.}

\section*{Acknowledgements}
Supported by the European Research Council (ERC FACTORY-CoG-6681839) and the ANR-3IA (ANITI).

\bibliographystyle{icml2020}
\bibliography{OrdNMF}

\begin{thebibliography}{31}
\providecommand{\natexlab}[1]{#1}
\providecommand{\url}[1]{\texttt{#1}}
\expandafter\ifx\csname urlstyle\endcsname\relax
  \providecommand{\doi}[1]{doi: #1}\else
  \providecommand{\doi}{doi: \begingroup \urlstyle{rm}\Url}\fi

\bibitem[Acharya et~al.(2015)Acharya, Ghosh, and
  Zhou]{acharya_nonparametric_2015}
Acharya, A., Ghosh, J., and Zhou, M.
\newblock Nonparametric {{Bayesian factor analysis}} for {{dynamic count
  matrices}}.
\newblock In \emph{Proc. {{International Conference}} on {{Artificial
  Intelligence}} and {{Statistics}} ({{AISTATS}})}, 2015.

\bibitem[Agresti \& Kateri(2011)Agresti and Kateri]{agresti_categorical_2011}
Agresti, A. and Kateri, M.
\newblock \emph{Categorical data analysis}.
\newblock {Springer}, 2011.

\bibitem[Ananth \& Kleinbaum(1997)Ananth and Kleinbaum]{ananth_regression_1997}
Ananth, C.~V. and Kleinbaum, D.~G.
\newblock Regression models for ordinal responses: A review of methods and
  applications.
\newblock \emph{International journal of epidemiology}, pp.\  1323--1333, 1997.

\bibitem[Basbug \& Engelhardt(2016)Basbug and
  Engelhardt]{basbug_hierarchical_2016}
Basbug, M.~E. and Engelhardt, B.~E.
\newblock Hierarchical {{compound Poisson factorization}}.
\newblock In \emph{Proc. {{International Conference}} on {{Machine Learning}}
  ({{ICML}})}, 2016.

\bibitem[{Bertin-Mahieux} et~al.(2011){Bertin-Mahieux}, Ellis, Whitman, and
  Lamere]{bertin-mahieux_million_2011}
{Bertin-Mahieux}, T., Ellis, D.~P., Whitman, B., and Lamere, P.
\newblock The {{million song dataset}}.
\newblock In \emph{Proc. {{International Society}} for {{Music Information
  Retrieval}} ({{ISMIR}})}, pp.\ ~10, 2011.

\bibitem[Canny(2004)]{canny_gap_2004}
Canny, J.
\newblock {{GaP}}: {{A factor model}} for {{discrete data}}.
\newblock In \emph{Proc. {{ACM International}} on {{Research}} and
  {{Development}} in {{Information Retrieval}} ({{SIGIR}})}, pp.\  122--129,
  2004.

\bibitem[Cemgil(2009)]{cemgil_bayesian_2009}
Cemgil, A.~T.
\newblock Bayesian {{inference}} for {{nonnegative matrix factorisation
  models}}.
\newblock \emph{Computational Intelligence and Neuroscience}, 2009.

\bibitem[Chu \& Ghahramani(2005)Chu and Ghahramani]{chu_gaussian_2005}
Chu, W. and Ghahramani, Z.
\newblock Gaussian processes for ordinal regression.
\newblock \emph{The Journal of Machine Learning Research}, pp.\  1019--1041,
  2005.

\bibitem[Consonni \& Marin(2007)Consonni and Marin]{consonni_meanfield_2007}
Consonni, G. and Marin, J.-M.
\newblock Mean-field variational approximate {{Bayesian}} inference for latent
  variable models.
\newblock \emph{Computational Statistics \& Data Analysis}, pp.\  790--798,
  2007.

\bibitem[F{\'e}votte \& Idier(2011)F{\'e}votte and
  Idier]{fevotte_algorithms_2011}
F{\'e}votte, C. and Idier, J.
\newblock Algorithms for nonnegative matrix factorization with the
  beta-divergence.
\newblock \emph{Neural computation}, pp.\  2421--2456, 2011.

\bibitem[Gopalan et~al.(2015)Gopalan, Hofman, and Blei]{gopalan_scalable_2015}
Gopalan, P., Hofman, J.~M., and Blei, D.~M.
\newblock Scalable recommendation with hierarchical {{Poisson}} factorization.
\newblock In \emph{Proc. {{Conference}} on {{Uncertainty}} in {{Artificial
  Intelligence}} ({{UAI}})}, pp.\  326--335, 2015.

\bibitem[Gouvert et~al.(2019)Gouvert, Oberlin, and
  F{\'e}votte]{gouvert_recommendation_2019}
Gouvert, O., Oberlin, T., and F{\'e}votte, C.
\newblock Recommendation from raw data with adaptive compound {{Poisson}}
  factorization.
\newblock In \emph{Proc. {{Conference}} on {{Uncertainty}} in {{Artificial
  Intelligence}} ({{UAI}})}, 2019.

\bibitem[Gutierrez et~al.(2015)Gutierrez, {Perez-Ortiz}, {Sanchez-Monedero},
  {Fernandez-Navarro}, and {Hervas-Martinez}]{gutierrez_ordinal_2015}
Gutierrez, P.~A., {Perez-Ortiz}, M., {Sanchez-Monedero}, J.,
  {Fernandez-Navarro}, F., and {Hervas-Martinez}, C.
\newblock Ordinal regression methods: Survey and experimental study.
\newblock \emph{IEEE Transactions on Knowledge and Data Engineering}, pp.\
  127--146, 2015.

\bibitem[Harper \& Konstan(2015)Harper and Konstan]{harper_movielens_2015}
Harper, F.~M. and Konstan, J.~A.
\newblock The movielens datasets: History and context.
\newblock \emph{ACM Transactions on Interactive Intelligent Systems (TIIS)},
  pp.\  1--19, 2015.

\bibitem[{Hernandez-Lobato} et~al.(2014){Hernandez-Lobato}, Houlsby, and
  Ghahramani]{hernandez-lobato_probabilistic_2014}
{Hernandez-Lobato}, J.~M., Houlsby, N., and Ghahramani, Z.
\newblock Probabilistic matrix factorization with non-random missing data.
\newblock In \emph{Proc. {{International Conference}} on {{Machine Learning}}
  ({{ICML}})}, pp.\  1512--1520, 2014.

\bibitem[Hu et~al.(2008)Hu, Koren, and Volinsky]{hu_collaborative_2008}
Hu, Y., Koren, Y., and Volinsky, C.
\newblock Collaborative filtering for implicit feedback datasets.
\newblock In \emph{Proc. {{IEEE International Conference}} on {{Data Mining}}
  ({{ICDM}})}, pp.\  263--272, 2008.

\bibitem[J{\"a}rvelin \& Kek{\"a}l{\"a}inen(2002)J{\"a}rvelin and
  Kek{\"a}l{\"a}inen]{jarvelin_cumulated_2002}
J{\"a}rvelin, K. and Kek{\"a}l{\"a}inen, J.
\newblock Cumulated gain-based evaluation of {{IR}} techniques.
\newblock \emph{ACM Transactions on Information Systems (TOIS)}, pp.\
  422--446, 2002.

\bibitem[Koren \& Sill(2011)Koren and Sill]{koren_ordrec_2011}
Koren, Y. and Sill, J.
\newblock {{OrdRec}}: An ordinal model for predicting personalized item rating
  distributions.
\newblock In \emph{Proc. {{ACM Conference}} on {{Recommender Systems}}
  ({{RecSys}})}, pp.\  117--124, 2011.

\bibitem[Koren et~al.(2009)Koren, Bell, and Volinsky]{koren_matrix_2009}
Koren, Y., Bell, R., and Volinsky, C.
\newblock Matrix factorization techniques for recommender systems.
\newblock \emph{Computer}, pp.\  30--37, 2009.

\bibitem[Lee \& Seung(1999)Lee and Seung]{lee_learning_1999}
Lee, D.~D. and Seung, H.~S.
\newblock Learning the parts of objects by non-negative matrix factorization.
\newblock \emph{Nature}, pp.\  788--791, 1999.

\bibitem[Lee \& Seung(2001)Lee and Seung]{lee_algorithms_2001}
Lee, D.~D. and Seung, H.~S.
\newblock Algorithms for non-negative matrix factorization.
\newblock In \emph{Advances in {{Neural Information Processing Systems}}
  ({{NIPS}})}, pp.\  556--562, 2001.

\bibitem[Liang et~al.(2016)Liang, Charlin, McInerney, and
  Blei]{liang_modeling_2016}
Liang, D., Charlin, L., McInerney, J., and Blei, D.~M.
\newblock Modeling user exposure in recommendation.
\newblock In \emph{Proc. {{International Conference}} on {{World Wide Web}}
  ({{WWW}})}, pp.\  951--961, 2016.

\bibitem[Lumbreras et~al.(2018)Lumbreras, Filstroff, and
  F{\'e}votte]{lumbreras_bayesian_2018}
Lumbreras, A., Filstroff, L., and F{\'e}votte, C.
\newblock Bayesian mean-parameterized nonnegative binary matrix factorization.
\newblock \emph{arXiv preprint arXiv:1812.06866}, 2018.

\bibitem[Ma et~al.(2011)Ma, Liu, King, and Lyu]{ma_probabilistic_2011}
Ma, H., Liu, C., King, I., and Lyu, M.~R.
\newblock Probabilistic factor models for web site recommendation.
\newblock In \emph{Proc. {{ACM International}} on {{Research}} and
  {{Development}} in {{Information Retrieval}} ({{SIGIR}})}, pp.\  265--274,
  2011.

\bibitem[McCullagh(1980)]{mccullagh_regression_1980}
McCullagh, P.
\newblock Regression models for ordinal data.
\newblock \emph{Journal of the Royal Statistical Society: Series B
  (Methodological)}, \penalty0 (2):\penalty0 109--127, 1980.

\bibitem[Paquet et~al.(2012)Paquet, Thomson, and
  Winther]{paquet_hierarchical_2012}
Paquet, U., Thomson, B., and Winther, O.
\newblock A hierarchical model for ordinal matrix factorization.
\newblock \emph{Statistics and Computing}, pp.\  945--957, 2012.

\bibitem[Rendle et~al.(2009)Rendle, Freudenthaler, Gantner, and
  Schmidt-Thieme]{BPR}
Rendle, S., Freudenthaler, C., Gantner, Z., and Schmidt-Thieme, L.
\newblock Bpr: Bayesian personalized ranking from implicit feedback.
\newblock In \emph{Proc. {{Conference}} on {{Uncertainty}} in {{Artificial
  Intelligence}} ({{UAI}})}, pp.\  452–461, 2009.

\bibitem[Stevens(1946)]{stevens_theory_1946}
Stevens, S.~S.
\newblock On the theory of scales of measurement.
\newblock 1946.

\bibitem[Verwaeren et~al.(2012)Verwaeren, Waegeman, and
  De~Baets]{verwaeren_learning_2012}
Verwaeren, J., Waegeman, W., and De~Baets, B.
\newblock Learning partial ordinal class memberships with kernel-based
  proportional odds models.
\newblock \emph{Computational Statistics \& Data Analysis}, pp.\  928--942,
  2012.

\bibitem[Walker \& Duncan(1967)Walker and Duncan]{walker_estimation_1967}
Walker, S.~H. and Duncan, D.~B.
\newblock Estimation of the probability of an event as a function of several
  independent variables.
\newblock \emph{Biometrika}, pp.\  167--179, 1967.

\bibitem[Zhou(2017)]{zhou_nonparametric_2017}
Zhou, M.
\newblock Nonparametric {{Bayesian}} negative binomial factor analysis.
\newblock \emph{Bayesian Analysis}, 2017.

\end{thebibliography}

\end{document}